\documentclass{article}

\usepackage[final]{neurips_2020}

\usepackage{natbib}
\usepackage{graphicx}
\usepackage{amsmath}
\usepackage{algorithm}
\usepackage{algorithmic}
\newcommand{\INDSTATE}[1][1]{\STATE\hspace{#1\algorithmicindent}}
\usepackage{caption}
\usepackage{subcaption}
\usepackage[colorinlistoftodos]{todonotes}

\title{Finding Useful Predictions by Meta-gradient Descent to Improve Decision-making}
\author{
  Alex Kearney \\
  Department of Computing Science\\
  University of Alberta\\
  Edmonton, AB, Canada \\
  \texttt{hi@alexkearney.com} \\
  \And
  Anna Koop \\
  Department of Computing Science\\
  University of Alberta\\
  Edmonton, AB, Canada \\
  \texttt{akoop@ualberta.ca} \\
  \And
  Johannes G\"unther\\
  Department of Computing Science \\
  University of Alberta\\
  \& \\
  Alberta Machine Intelligence Institute \\
  Edmonton, AB, Canada\\
  \texttt{gunther@ualberta.ca}
  \And
  Patrick M. Pilarski\\
  Department of Computing Science\\
  \& \\
 Department of Medicine\\
  University of Alberta\\
  Edmonton, AB, Canada \\
  \texttt{pilarski@ualberta.ca} \\
}
\date{2021}

\begin{document}
\maketitle
\begin{abstract}

  In computational reinforcement learning, a growing body of work seeks to express an agent's model of the world through predictions about future sensations. In this manuscript we focus on predictions expressed as General Value Functions: temporally extended estimates of the accumulation of a future signal. One challenge is determining from the infinitely many predictions that the agent could possibly make which might support decision-making. In this work, we contribute a meta-gradient descent method by which an agent can directly specify what predictions it learns, independent of designer instruction. 
  %We introduce a partially observable domain suited to this investigation, and  demonstrate that through interaction with the environment an agent can independently select predictions that resolve the partial-observability, resulting in performance similar to expertly chosen value functions.
  To that end, we introduce a partially observable domain suited to this investigation. We then demonstrate that through interaction with the environment an agent can independently select predictions that resolve the partial-observability, resulting in performance similar to expertly chosen value functions.
  By learning, rather than manually specifying these predictions, we enable the agent to identify useful predictions in a self-supervised manner, taking a step towards truly autonomous systems.
\end{abstract}
\section{Making Sense of The World Through Predictions}

It is often useful to break a challenging problem into sub-problems: progress on sub-tasks can support an agent's progress on a greater task, e.g., learning the values of states in order to approximate the optimal policy, or learning models of the world to better plan. One way an agent can create sub-problems and a world model is by learning predictions of its world---biological agents do this by building predictive sensorimotor models of their world \citep{rao1999predictive, wolpert1995internal,gilbert2009stumbling}. One principled and well understood way of making temporally extended predictions in reinforcement learning is by learning and maintaining value functions. Value functions predict the long-term expected accumulation of a signal in a given state \citep{sutton1988learning}, and can predict not only reward, but any signal available to an agent via its senses \citep{sutton2011horde}. Prior works have used general value estimates as features to adapt the control interfaces of bionic limbs \citep{edwards2016application}, design reflexive control systems for robots \citep{modayil2014prediction} and living cats \citep{dalrymple2020pavlovian}, and to inform industrial welding about the process quality \citep{GUNTHER20161}.

An open challenge when using GVFs is determining what to predict. Of all the possible predictions to make, which subset is most useful to inform and support decision making? This choice is typically made by the human designer of the system. However, previous work has used generate and test to choose which predictions are maintained, and which should be replaced \citep{schlegel2018baseline}. One hindrance of this method is the generator used to pick new predictions: the agent must explore a space of infinite predictions if they are chosen randomly. Moreover, common evaluation methods are not always reliable and can have adverse impacts on performance \citep{good-prediction}. Recent work has explored meta-gradient descent as a means of learning meta-parameters that specify the predictions \citep{DBLP:journals/corr/abs-1909-04607}; however, in this case the estimates were used as auxiliary tasks---the estimates themselves were not directly used in decision-making.

In this manuscript we propose a method of using meta-gradient descent to discover GVFs independent of human instruction and supervision. We do so by constructing a loss that shapes what the underlying predictions are about based on the control agent's learning process. All learning methods are updated incrementally and online. These value estimates can then be used directly as features by a control learner to solve a partially-observable problem.
%using a loss function that directly takes into account the control agent's performance.

%TODO -- work in they are independent of temporal span

\section{Learning What to Predict: An Architectural Proposal}

        \begin{figure}[b]
              \begin{subfigure}[b]{0.6\textwidth}
        \includegraphics[width=\linewidth,height=2in, keepaspectratio]{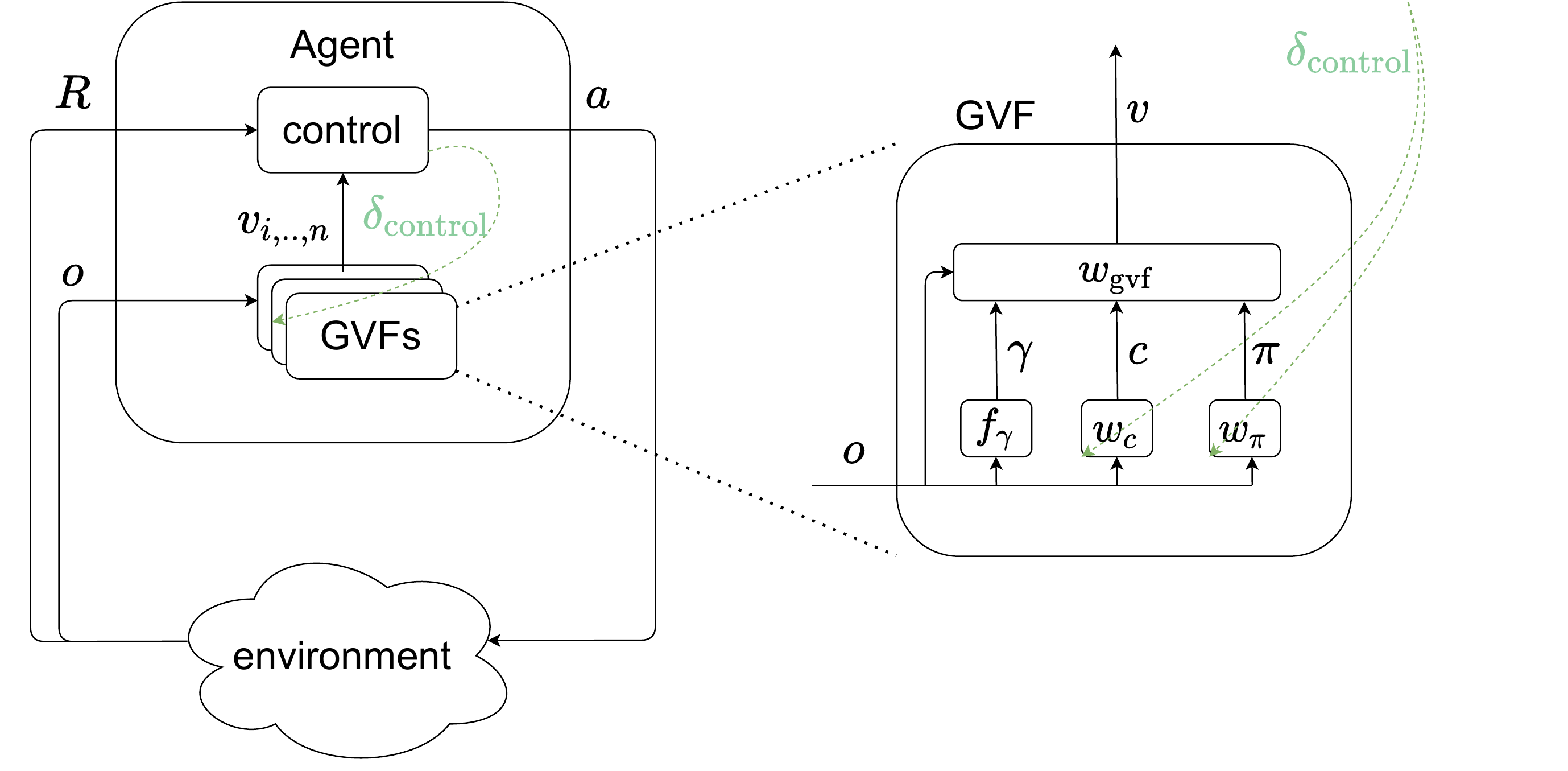}
          \centering
          \caption{A depiction of the agent-environment relationship showing how the agent processes information from the environment, and chooses an action. }
        \end{subfigure}
        \hfill
        \begin{subfigure}[b]{0.3\textwidth}
          \includegraphics[width=\linewidth,height=2.5in, keepaspectratio]{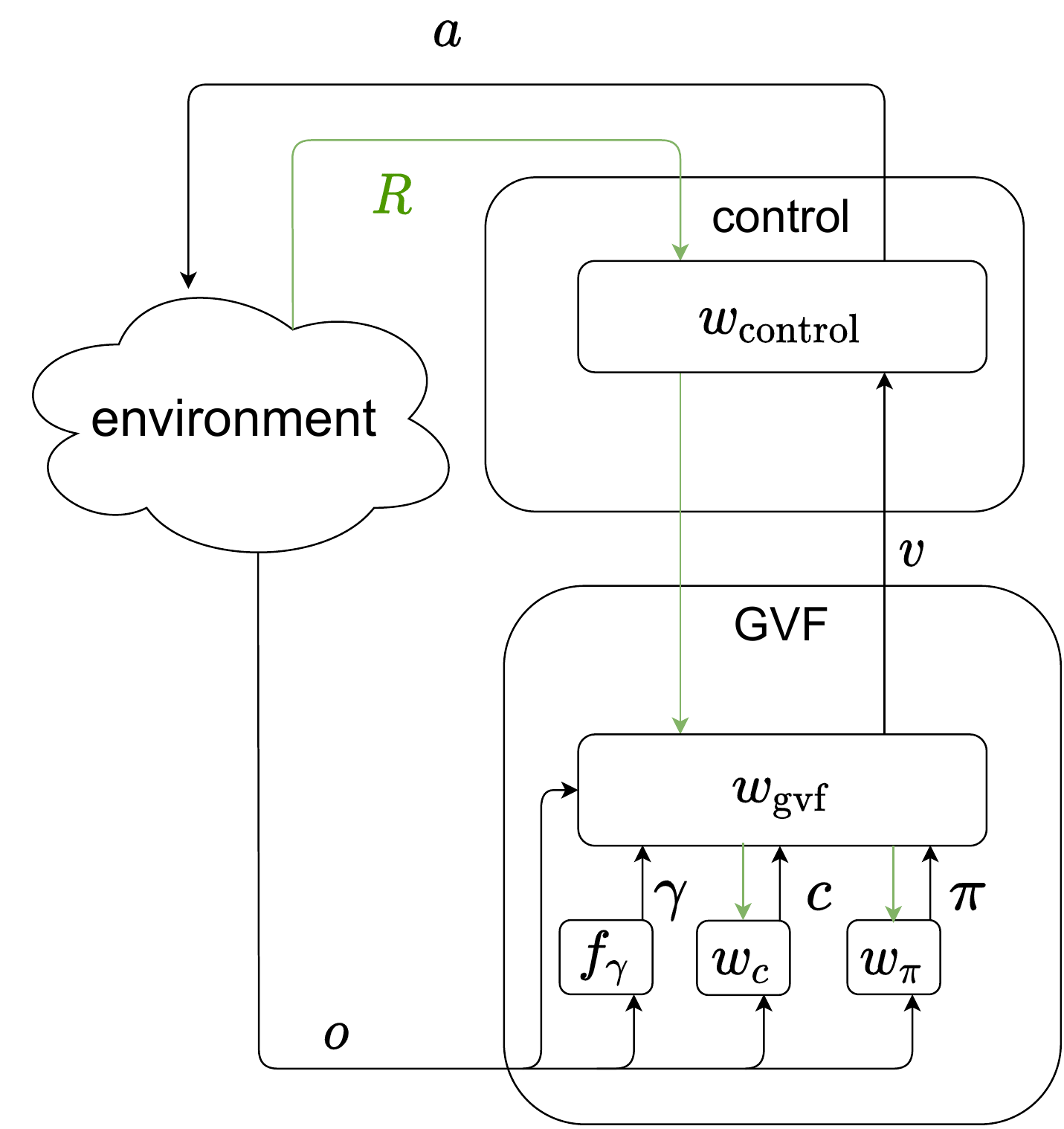}
          \centering
          \caption{A depiction of the indirect relationship between the error and the underlying weights.}
        \end{subfigure}
        \caption{Relationship between the sub-components of the agent and its environment. Denoted in green is the environments feedback in the form of reward/TD error.}
        \label{fig:architecture}
          \end{figure}

Our agent is structured in three parts (Figure \ref{fig:architecture}): 1) a control learner that chooses each action; 2) a collection of GVFs that approximate value-functions specified by some policies $\pi_{1 \cdots n}$, cumulants $c$, and discount functions $\gamma_{1 \cdots n}$; and 3) meta-weights that determine the policy $\pi$ ($w_{\pi}$) and cumulant $c$ ($w_{c}$) each GVF is conditioned on. This architecture is shown in Figure \ref{fig:architecture}; pseudocode describes the relationships between these components in Algorithm \ref{alg}.

On each time-step, the agent observes the current state of the environment $o_t$. A collection of $n$ GVFs perform a temporal difference update based on this observation, and prediction estimates are produced $v_{1\cdots n}$. These predictions $v_{1\cdots n}$ are given to the control agent as input features with which an action $a_{t+1}$ is chosen according to $\pi_{\text{control}}$. After taking an action, the agent observes a resulting reward $R_{t+1}$ and a following observation $o_{t+1}$ and the cycle repeats.

% {Johannes: @Patrick - is that enough for someone without extensive GVF knowledge to understand what is going on?

% TODO; you can mention that generate and test is only as good as the testing, and we're not quite sure how to do that (self-cite)

% TODO: the cumulant and why it matters---how it reflects the signal of interest.

Three parameters---the discount $\gamma$, policy $\pi$ and the cumulant $c$ determine what aspect of the environment each prediction is \textit{about}. The cumulant determines the signal of interest from the environment, and the policy defines what the agent is doing during the prediction. We define meta-weights $w_{\pi}$ and $w_{c}$ that determine the cumulant and policy a prediction is conditioned on. These meta-parameters are incrementally learned alongside the GVF they specify, affecting the GVF by determining how the values should change by modifying $w_{\text{gvf}}$ during each temporal difference update. 

% By changing the GVF estimates, these meta-weights indirectly change the control agent's features $v_{i \cdots n}$, thereby influencing how much reward the control learner can acquire. Using this indirect relationship, we can construct a gradient and modify the meta-weights $w_{\pi}$ and $w_{c}$ to learn predictions that better improve the control agent's policy.

There is an indirect relationship between the updates to the meta-weights, and the agent's TD error $ \delta_{\text{control}}$; through this relationship we can express a gradient that describes how the underlying weights $w_{pi}$ and $w_{c}$ which shape the meaning of a particular GVF influence the agent's Temporal Difference error: \(
  \frac{\partial \delta_{\text{control}}}{w_{\pi}} =
  \frac{\partial \delta_{\text{control}}}{\partial w_{\text{control}}}
  \frac{\partial w_{\text{control}}}{\partial w_{\text{gvf}}}
  \frac{\partial w_{\text{gvf}}}{\partial w_{\pi}}\). By this relationship, we can construct a loss function  $\mathcal{L}_{\pi}(w_{\text{control}},w_{\pi}) = \delta_{\text{control}}^{2}$ for the policy and the cumulant. Using meta-gradient descent, the underlying parameters $w_{c}$ and $w_{\pi}$ that output the cumulant $c$ and policy $\pi$ that a given agent is following perform meta-gradient descent can be updated as so:\(
w_{\pi} \gets w_{\pi} - \alpha_{\pi}  \nabla_{w_{\pi}} \mathcal{L}_{\pi}
\) and \(w_{c} \gets w_{c} - \alpha_{c}  \nabla_{w_{c}} \mathcal{L}_{c}
\).
    \begin{algorithm}[t]
    \caption{A meta-gradient approach to self-supervise prediction selection to inform control.}
          \begin{algorithmic}
            \INDSTATE[0] \textbf{INITALISE}: set control agent weights $w_\text{control}$, GVF weights $w_{\text{gvf}, i \cdots n}$, and meta weights $w_{c, i}$, $w_{\pi, i}$. Choose activations $\phi_\text{cumulant}$ and $\phi_\text{policy}$. Choose step-size $\alpha$ for the control agent, GVFs, and meta-parameters independently. Set an L2 $\lambda$. Set an $\epsilon$ e-greedy for action-selection.
            \INDSTATE[0] \textbf{START: } Make initial observation $o_{0}$, take initial action $a_{0}$. The $\text{gvf-state}_0$ is $o_{0}$, $a_{0}$. \INDSTATE[0] Produce GVF estimates $V_{0}$; these estimates form $\text{control-state}_{0}$.
            \INDSTATE[1] \textbf{for} t=1, to final time-step T:
            \INDSTATE[2] With probability $\epsilon$ select a random action $a_{t}$.
            \INDSTATE[2] Else select action $a_{t} = argmax_{a} Q(\phi(v_{t}), a_{t})$.
            \INDSTATE[2] Observe $o_{t+1}$ and $r_{t+1}$ resulting from action $a_{t}$.
            \INDSTATE[2] \textbf{Perform meta-gradient descent}
            \INDSTATE[4] Take gradient steps on $\mathcal{L}$ with respect to both $w_{\text{policy}}$ and $w_{\text{cumulant}}$.
            \INDSTATE[2] \textbf{Update GVFs}
            \INDSTATE[4] Output current cumulant $c = \phi_{\text{cumulant}}(o_{t}, w_{\text{cumulant}})$.
            \INDSTATE[4] Output current policy $\pi_{\text{gvf}} = \phi_{\text{policy}}(o_{t},w_{\text{policy}})$.

            \INDSTATE[4] Approximate state for GVFs: $\text{gvf-state}=o_{t},a_{t},v_{t}$.
            \INDSTATE[4] Update each GVF's parameters $w_{\text{gvf}}$ given their computed $\pi$, $c$, and fixed $\gamma$.
            \INDSTATE[2] \textbf{Update control policy}
            \INDSTATE[4] Output current estimate for each GVF given $v_{t+1}(\text{gvf-state}_{t}, w_{\text{gvf}})$.
            \INDSTATE[4] Approximate state for control learner:
            $\text{control-state}_{t+1} = v_{t+1}$.
            \INDSTATE[4] Update control agent's Q values given $\text{control-state}_{t}$, $a_{t}$, $r_{t+1}$,
            \INDSTATE[4] And $\text{control-state}_{t+1}$.
            \INDSTATE[1]\textbf{end}
          \end{algorithmic}
          \label{alg}
        \end{algorithm}
     
\section{Monsoon World: A Partially Observable Environment}
\label{experiment-description}

\begin{figure}[t]
\centering
\begin{subfigure}[b]{0.4\textwidth}
\includegraphics[width=\linewidth,height=4.5cm,keepaspectratio]{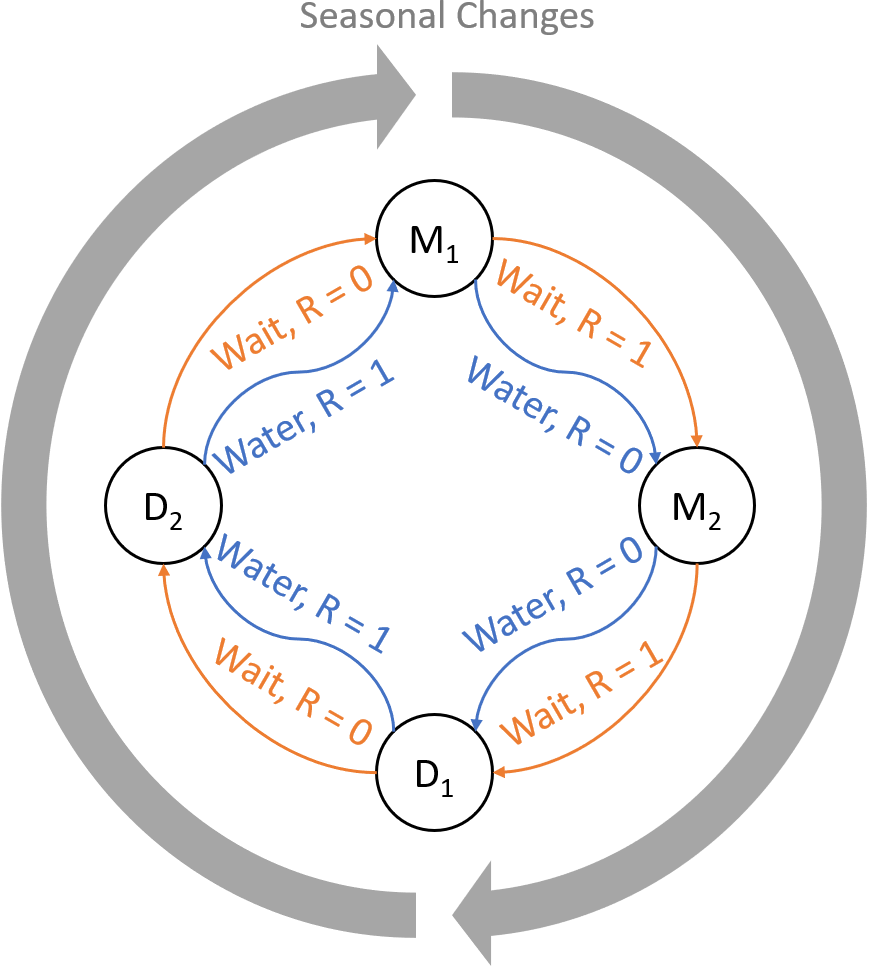}
\centering
\caption{There are four states: two monsoon and two drought (inner circles). The outer arrows indicate how the seasons change as the agent transitions through the cycle.}
\label{monsoon}
\end{subfigure}\hfill
\begin{subfigure}[b]{0.59\textwidth}
\includegraphics[width=1\linewidth, height=4.5cm, keepaspectratio]{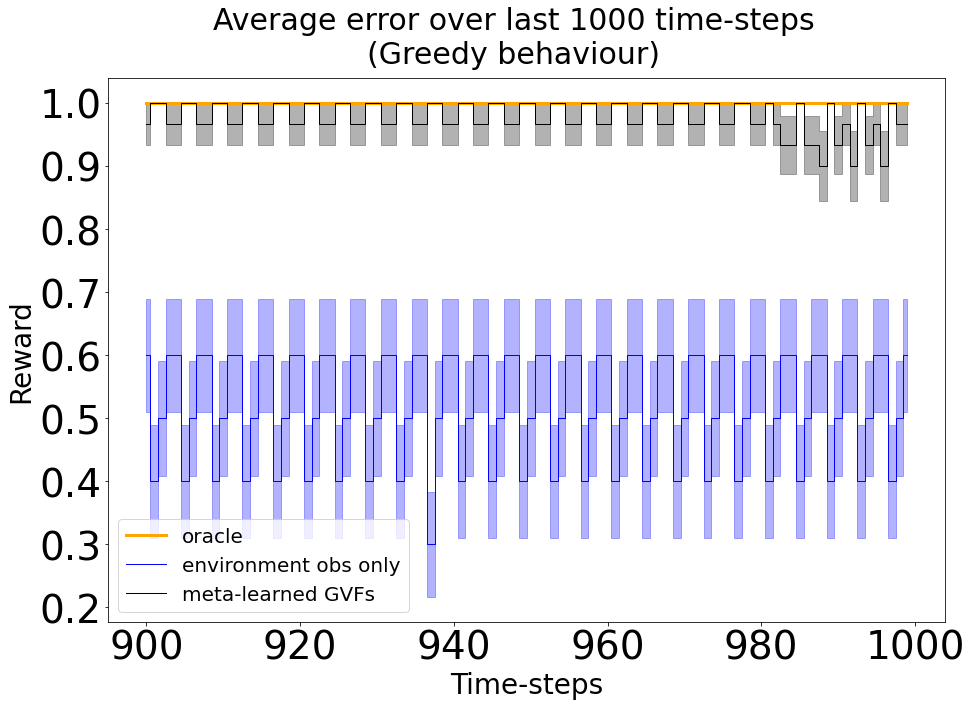}
\centering
\caption{Three different learners that use 1) the environmental observations as inputs (blue),  2) two additional predictions that are known to express the seasons (in orange), 3) two additional predictions that are updated using meta-gradient descent (in black).}
\label{results}
\end{subfigure}
\centering
\caption{Monsoon World and comparison of learned policies. Each independent agent in \ref{results} is averaged over 30 independent trials. Error bars are standard error.}
\label{intro}

\end{figure}

We evaluate meta-gradient discovery of GVFs using a partially observable control problem, Monsoon World (Figure \ref{monsoon}). In Monsoon World, there are two seasons: monsoon and drought. The agent tends to a field by choosing to either water, or not water their farm. Watering the field during a drought will result in a reward of 1; watering the field during monsoon season does not produce growth and results in a reward of 0, and vice versa during a monsoon. If the agent chooses the right action corresponding to the underlying season, a reward of 1 can be obtained on each time-step. Regardless of the action chosen by the agent, time progresses.

In this environment the agent cannot observe the underlying season that determines the outcome of their action. While the agent cannot directly observe seasons, it can observe something impacted by the seasons--the result of a given action. 

This monsoon problem can be solved, and an optimal policy found, if the agent reliably estimates how long until watering produces a particular result. This can be done by learning \textit{echo GVFs} \citep{schlegel2021general}. Echo GVFs estimate the time to an event using a state-conditioned discount and cumulant. In this case, estimating how long until there is growth $o_{i,t}=1$, or there is no growth $o_{i,t}=0$ when watering: \(c = 1 \quad \text{if} \quad o_{i,t} = 1 \text{; else} \quad c = 0 \).  Similarly, a state-dependent discounting function terminates the accumulation $\gamma(s_t,a_t,s_{t+1})$, where \(\gamma = 0 \quad \text{if} \quad c = 1 \text{; else} \quad 0.9\). These estimates can be learnt off-policy using a deterministic policy (e.g., ``if the agent waters'' \(\pi = [0,1]\)).

Having constructed the aforementioned GVFs, we are now able to express what is hidden from our observation stream: how long until the next season. While no information was given about the season, by relating what is sensed by the agent with the actions that were taken by the agent, we are able to learn about the seasons indirectly.

\section{Discovering GVFs in Monsoon World}

We now answer the question: ``Can an agent find useful predictions by performing meta-gradient descent?'' To do so, we compare three different agent configurations (Figure \ref{results}): 1) a baseline agent that only receives environmental observations as inputs, 2) an agent that in addition to the environmental observations, receives the estimates of two GVFs with cumulants and policies known to be effective in capturing the underlying seasons, and 3) an agent that has two additional predictions that are learned through meta-gradient descent.

When GVFs are specified via meta-gradient descent, we initialise policies  to an equiprobable weighting of actions and cumulants to an equal weighting of observations. The policy weights are passed through a Softmax activation function so that their sum is between 1 and 0, and the cumulants are passed through a sigmoid activation to bound the cumulant between [0,1].  The meta-weights are updated each time-step incrementally. We apply L2 to the loss with $\lambda = 0.001$. Additional details are in Appendix \ref{appendix}.

As introduced in Section  \ref{experiment-description}, observations alone are insufficient to determine the optimal action on a given time-step. The policy learnt using only environment observations is roughly equivalent to equiprobably choosing an action: the learned policy is no better than a coin-toss (Figure \ref{results}, depicted in blue). When expertly specified estimates are learned and provided as inputs in addition to the environmental observations (orange). the learned policy is approximately optimal: using predictions that estimate the time to each season optimal actions are taken most of the time. By using meta-gradient descent, the agent was able to select its own predictive features without any prior knowledge of the domain. {\em Using Meta-gradient descent, the agent is able to solve the task with performance on-par with the hand-crafted solution without being given what to predict.}

% \begin{figure}
% \centering
% \begin{subfigure}[b]{0.48\textwidth}
% \includegraphics[width=\linewidth, keepaspectratio]{figures/fail-vs-success.png}
% \centering
% \caption{Change in the target policy $\pi$ for both GVFs. There are two available actions, and the Each plot has 30 independent trials.}
% \end{subfigure}\hfill
% \begin{subfigure}[b]{0.48\textwidth}
% \includegraphics[width=\linewidth,keepaspectratio]{figures/cumulants.png}
% \centering
% \caption{Change in the target policy $\pi$ for both GVFs. There are two available actions, and the Each plot has 30 independent trials.}
% \end{subfigure}
% \caption{How the meta-parameters vary during the experiment.}
% \end{figure}

\section{Limitations \& Future Work }
We introduced a new approach to meta-learning predictions where GVFs outputs are used as features by a control agent. We found that an agent with no prior knowledge of the environment was able to select predictions that yielded performance equitable to agents using expertly chosen predictive features. This success opens up several interesting questions for future work: 1) How well does meta-gradient selection perform in domains with higher-dimensional observations and more actions? 2) How well would meta-gradient GVF selection perform in non-stationary domains? Finally, our proposed approach is sensitive to the meta step-sizes that govern the incremental updates. How optimisers or step-size adaptations could improve robustness has yet to be explored.

\section{Conclusion}
In this paper we demonstrated how predictions in the form of GVFs can be decided upon and learned by meta-gradient descent alongside a policy for agent action selection. Doing so, we enable our agent to learn about its environment in a self-supervised and independent way. We evaluate our approach on a partially-observable MDP, called Monsoon world. Our results demonstrate that an agent can independently specify GVFs that enable performance comparable to expertly chosen predictions that remove the partial-observability. This work therefore tackles one of the most important problems in prediction-based self-supervised learning.

% \bibliographystyle{apa.bst}
% \bibliography{bib.bib}

\begin{thebibliography}{}

\bibitem[\protect\astroncite{Dalrymple et~al.}{2020}]{dalrymple2020pavlovian}
Dalrymple, A.~N., Roszko, D.~A., Sutton, R.~S., and Mushahwar, V.~K. (2020).
\newblock Pavlovian control of intraspinal microstimulation to produce
  over-ground walking.
\newblock {\em Journal of neural engineering}, 17(3):036002.

\bibitem[\protect\astroncite{Edwards et~al.}{2016}]{edwards2016application}
Edwards, A.~L., Dawson, M.~R., Hebert, J.~S., Sherstan, C., Sutton, R.~S.,
  Chan, K.~M., and Pilarski, P.~M. (2016).
\newblock Application of real-time machine learning to myoelectric prosthesis
  control: A case series in adaptive switching.
\newblock {\em Prosthetics and orthotics international}, 40(5):573--581.

\bibitem[\protect\astroncite{Gilbert}{2009}]{gilbert2009stumbling}
Gilbert, D. (2009).
\newblock {\em Stumbling on happiness}.
\newblock Vintage Canada.

\bibitem[\protect\astroncite{Günther et~al.}{2016}]{GUNTHER20161}
Günther, J., Pilarski, P.~M., Helfrich, G., Shen, H., and Diepold, K. (2016).
\newblock Intelligent laser welding through representation, prediction, and
  control learning: An architecture with deep neural networks and reinforcement
  learning.
\newblock {\em Mechatronics}, 34:1--11.
\newblock System-Integrated Intelligence: New Challenges for Product and
  Production Engineering.

\bibitem[\protect\astroncite{Kearney et~al.}{2021}]{good-prediction}
Kearney, A., Koop, A., and Pilarski, P.~M. (2021).
\newblock What's a good prediction? {Issues} in evaluating general value
  functions through error.

\bibitem[\protect\astroncite{Modayil and Sutton}{2014}]{modayil2014prediction}
Modayil, J. and Sutton, R.~S. (2014).
\newblock Prediction driven behavior: Learning predictions that drive fixed
  responses.
\newblock In {\em Workshops at the Twenty-Eighth AAAI Conference on Artificial
  Intelligence}.

\bibitem[\protect\astroncite{Rao and Ballard}{1999}]{rao1999predictive}
Rao, R.~P. and Ballard, D.~H. (1999).
\newblock Predictive coding in the visual cortex: {A} functional interpretation
  of some extra-classical receptive-field effects.
\newblock {\em Nature neuroscience}, 2(1):79--87.

\bibitem[\protect\astroncite{Schlegel et~al.}{2021}]{schlegel2021general}
Schlegel, M., Jacobsen, A., Abbas, Z., Patterson, A., White, A., and White, M.
  (2021).
\newblock General value function networks.
\newblock {\em Journal of Artificial Intelligence Research}, 70:497--543.

\bibitem[\protect\astroncite{Schlegel et~al.}{2018}]{schlegel2018baseline}
Schlegel, M., White, A., and White, M. (2018).
\newblock A baseline of discovery for general value function networks under
  partial observability.
\newblock In {\em NeurIPS Workshop on Reinforcement Learning under Partial
  Observability): Montreal, Canada}.

\bibitem[\protect\astroncite{Sutton}{1988}]{sutton1988learning}
Sutton, R.~S. (1988).
\newblock Learning to predict by the methods of temporal differences.
\newblock {\em Machine learning}, 3(1):9--44.

\bibitem[\protect\astroncite{Sutton et~al.}{2011}]{sutton2011horde}
Sutton, R.~S., Modayil, J., Delp, M., Degris, T., Pilarski, P.~M., White, A.,
  and Precup, D. (2011).
\newblock Horde: A scalable real-time architecture for learning knowledge from
  unsupervised sensorimotor interaction.
\newblock In {\em The 10th International Conference on Autonomous Agents and
  Multiagent Systems-Volume 2}, pages 761--768.

\bibitem[\protect\astroncite{Veeriah
  et~al.}{2019}]{DBLP:journals/corr/abs-1909-04607}
Veeriah, V., Hessel, M., Xu, Z., Lewis, R.~L., Rajendran, J., Oh, J., van
  Hasselt, H., Silver, D., and Singh, S. (2019).
\newblock Discovery of useful questions as auxiliary tasks.
\newblock {\em CoRR}, abs/1909.04607.

\bibitem[\protect\astroncite{Wolpert et~al.}{1995}]{wolpert1995internal}
Wolpert, D.~M., Ghahramani, Z., and Jordan, M.~I. (1995).
\newblock An internal model for sensorimotor integration.
\newblock {\em Science}, 269(5232):1880--1882.

\end{thebibliography}

\newpage
\appendix
\section{Experiment Details}
\label{appendix}

Experiments ran for a total of one million time-steps. Each agent had a training phase of 990,000 time-steps. The final 1000 time-steps the agent's performance is evaluated: $\epsilon$ is set to 0 and actions are chosen greedily so that we can compare average reward given the learned policies.

\subsection{Function Approximators}

We use different function approximators to transform the given inputs to an \textit{agent-state} $s_t = \phi(o_t,v_t)$. Echo GVFs are in log-space; before using them as inputs, we apply a transformation to them as follows:

    \begin{algorithm}[H]
    \caption{Log-transform of prediction estimates}
          \begin{algorithmic}
          \INDSTATE[0]\# Where $v$ is are the value estimate from $n$ GVFs. 
            \INDSTATE[0]$\text{transform}(v):$
            \INDSTATE[1]$v \gets \text{clip}(log(v)/log(0.9), 0,1)$
            \INDSTATE[1]return $v$
          \end{algorithmic}
          \label{FA}
        \end{algorithm}

We use state aggregation to transform the estimates produced by each GVF into a binary feature vector $s_t$ such that the value.

    \begin{algorithm}[H]
    \caption{State aggregation of predictions}
          \begin{algorithmic}
          \INDSTATE[0]\# Where $v$ is are the value estimate from $n$ GVFs. 
          \INDSTATE[0]\# Where memsize is the allocated length for the binary feature vector.
            \INDSTATE[0]$\text{state}(v, \text{memsize}):$
            \INDSTATE[1]$s = \text{zeros}(\text{memsize})$
            \INDSTATE[1]$i \gets v[0] + v[1]*10$ \quad \# this assumes that each $v_i < 10$
            \INDSTATE[1]$s[i] = 1$
            \INDSTATE[1]return $s$
          \end{algorithmic}
          \label{FA}
        \end{algorithm}

The function approximation for each agent is as follows:

\textbf{Environment Observations Only}

\begin{enumerate}
    \item Control Agent: state aggregation.
    \item GVFs: state aggregation.
    \item Meta-parameters: n/a.
\end{enumerate}

\textbf{Expert Chosen Predictions \& Environment Observations}

\begin{enumerate}
    \item Control Agent: state aggregation.
    \item GVFs: state aggregation.
    \item Meta-parameters: n/a.
\end{enumerate}

\textbf{Meta-gradient Learned Predictions \& Environment Observations}

\begin{enumerate}
    \item Control Agent: no function approximator; a linear combination of weights and inputs.
    \item GVFs: state aggregation.
    \item Meta-parameters: no function approximator; a linear combination of weights and inputs.
\end{enumerate}

\subsection{Parameter Settings}

Parameters were chosen by performing a sweep across different values, choosing the best performing combination for each agent.

\begin{table}[H]
    \centering
    \begin{tabular}{| c | c | c | c | c | c | c |}
         \hline
         Agent configuration & $\epsilon$ & $\alpha_{control}$ & $\alpha_{gvfs}$ & $\alpha_{\pi}$ & $\alpha_{c}$ \\
         \hline 
         Environment Obs Only & 0.1 & 0.01 & 0.1 & n/a & n/a \\ \hline
         Expert Chosen Predictions &  0.1 & 0.01 & 0.1 & n/a & n/a \\ \hline
         Meta-gradient Learned Predictions &  0.5 & 0.0001 & 0.1 & 0.001 & 0.1 \\ \hline
    \end{tabular}
    \caption{Parameter settings for different agent configurations}
    \label{tab:my_label}
\end{table}

\subsection{Meta-parameter Specification}

The policy $\pi$ is a deterministic policy. The meta-weights determine the policy a GVF is conditioned on, but they are not a function of the observations: $\pi \gets \text{softmax}(w_\pi)$.

The cumulant $c$ is a function of the observations such that $\text{sigmoid}(w_c^\top o_t)$, where $w_c$ are the meta-weights for the cumulant, and $o_t$ is the present environment observation.
\end{document}